\documentclass[sigconf,balance=false]{acmart}

\AtBeginDocument{%
  \providecommand\BibTeX{{%
    \normalfont B\kern-0.5em{\scshape i\kern-0.25em b}\kern-0.8em\TeX}}}

\setcopyright{acmcopyright}
\copyrightyear{2018}
\acmYear{2018}
\acmDOI{XXXXXXX.XXXXXXX}

\acmConference[Conference acronym 'XX]{Make sure to enter the correct
  conference title from your rights confirmation emai}{June 03--05,
  2018}{Woodstock, NY}
\acmPrice{15.00}
\acmISBN{978-1-4503-XXXX-X/18/06}

\begin{document}

\title{Decision Models for Selecting Architectural Patterns for Federated Machine Learning Systems}

\author{Sin Kit Lo}
\email{Kit.Lo@data61.csiro.au}
\affiliation{%
  \institution{CSIRO Data61 and University of New South Wales}
  \country{Australia}
}
\author{Qinghua Lu}
\email{qinghua.lu@data61.csiro.au}
\affiliation{%
  \institution{CSIRO Data61}
  \country{Australia}
}

\author{Hye-young Paik}
\email{h.paik@unsw.edu.au}
\affiliation{%
  \institution{University of New South Wales}
  \country{Australia}
}

\author{Liming Zhu}
\email{Liming.Zhu@data61.csiro.au}
\affiliation{%
  \institution{CSIRO Data61}
  \country{Australia}
}

\renewcommand{\shortauthors}{Lo and Lu, et al.}

\begin{abstract}
  Federated machine learning is growing fast in academia and industries as a solution to solve data hungriness and privacy issues in machine learning. Being a widely distributed system, federated machine learning requires various system design thinking. To better design a federated machine learning system, researchers have introduced multiple patterns and tactics that cover various system design aspects. However, the multitude of patterns leaves the designers confused about when and which pattern to adopt. In this paper, we present a set of decision models for the selection of patterns for federated machine learning architecture design based on a systematic literature review on federated machine learning, to assist designers and architects who have limited knowledge of federated machine learning. Each decision model maps functional and non-functional requirements of federated machine learning systems to a set of patterns. We also clarify the drawbacks of the patterns. We evaluated the decision models by mapping the decision patterns to concrete federated machine learning architectures by big tech firms to assess the models' correctness and usefulness. The evaluation results indicate that the proposed decision models are able to bring structure to the federated machine learning architecture design process and help explicitly articulate the design rationale.
\end{abstract}





\maketitle

\section{Introduction}
The growth of the idea of industry 4.0 and cloud computing resulted in the exponential increase in data dimensions and the availability of data to generate useful insights~\cite{s19204354}. The overall increase in data and computation capability of computers accelerated the adoption of machine learning for data analysis in multiple areas. However, many machine learning systems suffer from insufficient training data due to data privacy concerns. Data privacy as an important ethical principle of machine learning systems \cite{jobin2019global} induced the regularisation of access to privacy-sensitive data. Furthermore, trustworthy AI has become an emerging topic lately due to the new ethical, legal, social, and technological challenges brought on by the technology~\cite{TAI2020}.

Google introduced federated machine learning as a new concept for distributed machine learning settings in 2017~\cite{mcmahan2017communicationefficient}. The settings utilize a central server to orchestrate machine learning model training on widely distributed devices using their locally collected data, without central collection and preprocessing of training data. Hence, federated machine learning is able to solve data-sharing, privacy, and resource-sharing restriction challenges. However, a federated machine learning system presents more architectural design challenges~\cite{10.1145/3450288,10.1145/3298981}, especially when dealing with the interactions between the central server and client devices to manage the drawbacks amongst the software quality attributes. For instance, a federated machine learning system faces architecture challenges such as the need to consider how to actively manage multiple client devices while preventing malicious participants, or how to resolve the statistical and system heterogeneity across the client devices to maintain the model training performance~\cite{10.1145/3298981}. Various federated machine learning software architectural challenges and propose approaches to tackle the challenges were articulated and presented in our systematic literature review (SLR)~\cite{10.1145/3450288}. We have also summarised a set of software architectural patterns~\cite{lo2021architectural} to address the different requirements from different research articles and industrial practices. Despite having various patterns and solutions, architects may find it difficult to choose when and how to use them. Hence, we aim to structure the patterns and solutions to assist architects in selecting appropriate patterns during the federated machine learning system design through a series of pattern selection decision models. The goal is to provide guidance for federated machine learning architecture design decisions that meet the intended requirements while taking drawbacks and constraints into consideration.

The remainder of the paper is organised as follows: Section~\ref{fl_intro} gives an explanation of federated machine learning. Section~\ref{DM_overview} presents the notation and overview of the proposed decision models and elaborates on the 4 decision models for different aspects of architecture design. The evaluation of the decision models is presented in Section~\ref{evaluation}. Section~\ref{related_work} covers the related work on decision models, machine learning, and federated machine learning patterns. Finally, Section~\ref{conclusion} concludes the paper.

\section{Federated Machine Learning: Next-word Prediction}\label{fl_intro}

\begin{figure}
    \centering
    \includegraphics[width=\linewidth]{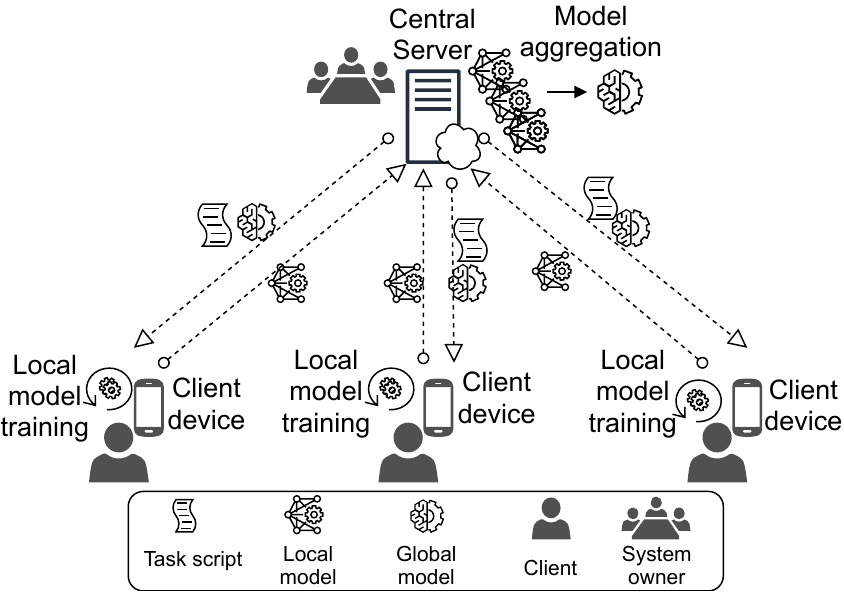}
    \caption{Federated machine learning overview~\cite{lo2021architectural}
    \label{Fig:FLOps}}
\end{figure}

{Fig.~\ref{Fig:FLOps} illustrates the overview of federated machine learning, under a next-word prediction example in a mobile phone keyboard application. There are two types of system nodes: (1) central server, and (2) client device.  Firstly, the learning coordinator (central server) of a federated machine learning system initiates the keyboard application. The contributor clients (client devices) are mobile phone users. The federated machine learning process begins with the creation of a training task (includes training hyperparameters, epochs, aggregation rounds, deployment strategy, etc.), usually by the central server. For instance, the keyboard application is embedded with an initial global model (including scripts \& hyperparameters), which is then sent to the participating client devices.}

On the client side, the global model is received and the training is performed locally across the client devices using the raw client data. In each training round, the client device will perform one for several training epochs to optimise the local model. In this case, the smartphones optimise the model using the keyboard typing data. After that, the updated local model is submitted by each participating client device to the central server to perform model aggregation to form a new version of the global model. The new global model is re-distributed to the client devices for the next aggregation round. The entire process repeats until the global model converges. After the completion of training, the central server deploys the converged global model to the client devices. In this example, the keyboard application provider applies the converged model in the latest version of the application for existing or new application users to perform the next word predictions. 

As the entire process repeats, communication and computation costs will be high. For instance, communication bandwidth is highly consumed by multiple client devices when they communicate with the central server and it increases with the scale of the number of devices connected. Furthermore, client devices may run on different operating systems and have diverse communication and computation resources, which trigger the \textit{system heterogeneity} challenges. One example is the difference in operating systems and computation resources trigger interoperability and model consistency issues. Similarly, the \textit{statistical heterogeneity} issue also exists, caused by the difference in data distributions across all client devices. The system is also troubled by \textit{system reliability} issue with the possibility of adversarial nodes participating in the training process, poisoning the system and the model quality and the central server is being exposed as the~\textit{single-point-of-failure}. Federated machine learning generates multiple versions of the local and global models created that need to be managed. However, the \textit{traceability} of the system is challenging as model provenance for all the local and global models is difficult as the system scale up. For example, local models are trained by the large scale of privately owned devices, using data that are unseen and not processed by the central server. It is challenging to know which version of local devices are trained with which version of local data has been aggregated into which version of a global model. Finally, due to the limited resources available on each device, the motivation of the client to join the federated machine learning process becomes weak, which induces the \textit{client motivatability} challenge. Client device owners might not want to use smartphones to train local models without reward as the process consumes battery life, communication bandwidth, and computation resources.

\begin{figure*}
    \centering
    \includegraphics[width=0.6\linewidth]{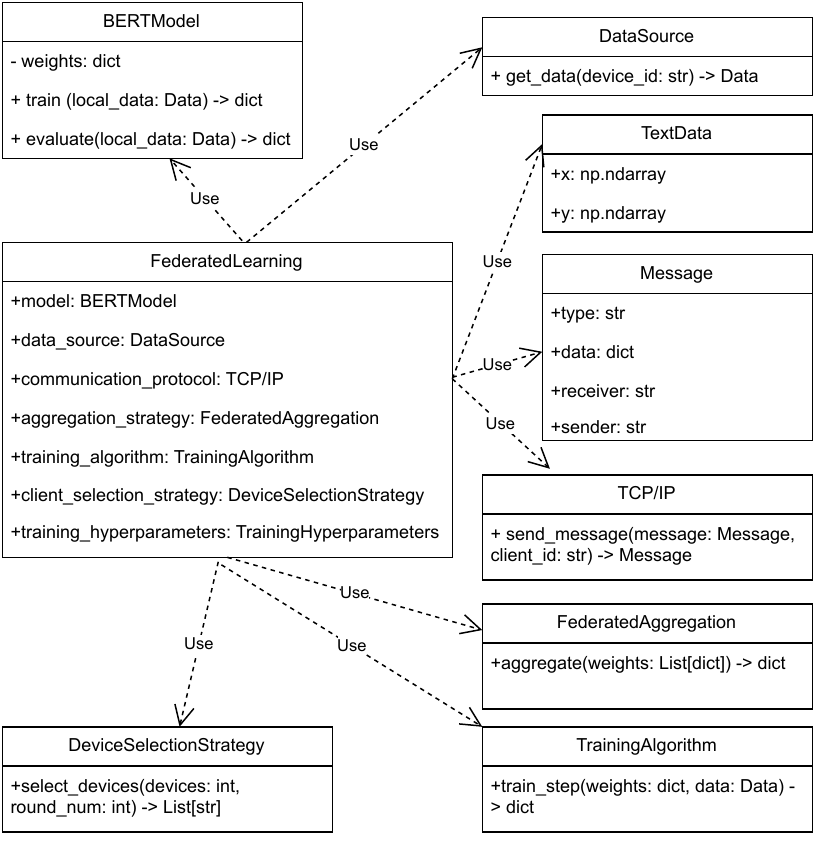}
    \caption{Federated machine learning Domain Model}
    \label{fig:uml}
\end{figure*}

Fig.~\ref{fig:uml} presents a domain model for federated machine learning systems. The dashed line specifically indicates that the association between \textbf{`FederatedLearning'} and the other classes is a ``use'' relationship. This means that \textbf{`FederatedLearning'} applies instances of the other classes to carry out its functionality. In this diagram, \textbf{`FederatedLearning'} represents the main class for the federated machine learning system. For example, \textbf{`FederatedLearning'} uses an instance of \textbf{`BERTModel'} to train the BERT (Bidirectional Encoder Representations from Transformers)~\cite{devlin2019bert} model used in the federated machine learning system. Similarly, \textbf{`FederatedLearning'} uses an instance of \textbf{`TCP/IP'} to handle the communication between the devices and the server in the federated machine learning system. It contains attributes and methods for various components of the system, such as the model, data source, communication protocol, aggregation strategy, training algorithm, and client selection strategy. \textbf{`BERTModel'} represents the machine learning model used in the federated machine learning system. It has attributes for the weights of the model and methods for training and evaluation. The \textbf{`DataSource'} represents the source of data for the federated machine learning system. It has a method for getting the data for a specific client. \textbf{`TCP/IP'} represents the method of communication between the clients and the server in the federated machine learning system. It has a method for sending messages between clients and the server. The \textbf{`FederatedAveraging'} represents the method for aggregating the weights of the models from the clients in the federated machine learning system while the \textbf{`TrainingAlgorithm'} represents the algorithm used for training the machine learning model in the federated machine learning system. The \textbf{`DeviceSelectionStrategy'} represents the method for selecting the clients to participate in the federated machine learning system. It has a method for selecting clients. Lastly, \textbf{`TextData'} represents the data used for training and evaluation in the federated machine learning system.

\section{Decision Models} \label{DM_overview}

\begin{figure*}
    \centering
    \includegraphics[width=0.6\linewidth]{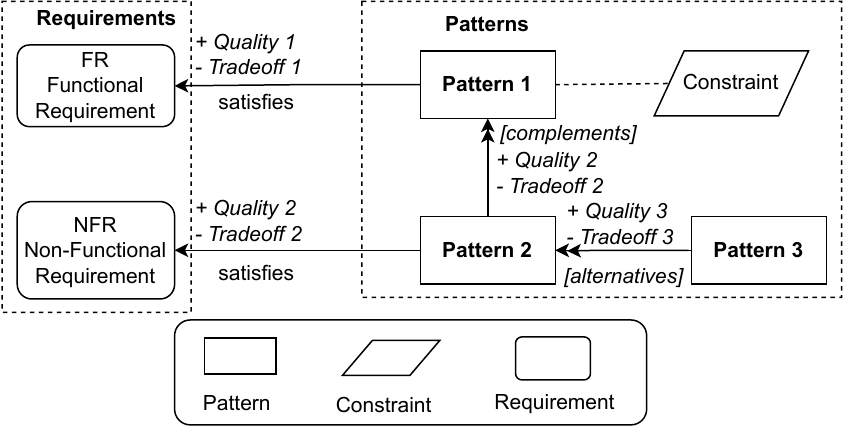}
    \caption{Decision model notations~\cite{7516811}}
    \label{fig:Decision model notation}
\end{figure*}

\begin{table*}[tbp]
\footnotesize
\centering
\caption{Overview of architectural patterns for federated machine learning~\cite{lo2021architectural}}
\label{tab:overview}
\begin{tabular}{p{0.2\columnwidth}p{0.3\columnwidth}p{1.3\columnwidth}}
\toprule

\textbf{\footnotesize{Category}} &
\textbf{\footnotesize{Name}} &
\textbf{\footnotesize{Summary}}\\
\midrule

\textbf{Federation management patterns} & {Client registry} & Exposes addresses and training capabilities of machine learning devices to the central server and maintains the information of all the participating client devices for federation management.\\
\cmidrule(l){2-3}

& {Client selector} & Actively selects the client devices for a certain round of training according to the predefined criteria to increase model quality and computation efficiency.\\
\cmidrule(l){2-3}

& {Client cluster} & Groups the client devices (i.e., model trainers) based on their similarity of certain characteristics (e.g., available resources, data distribution, features, geolocation) to increase the model quality and training efficiency.\\

\cmidrule(l){1-3}
\textbf{Model management \& configuration patterns} & {Message compressor} & Reduces message (global and local models) data size through different ways (compression, pruning, etc.) before every round of model exchange to increase communication efficiency.\\

\cmidrule(l){2-3}
& {Model co-versioning registry} & Stores and aligns the local models from each client with the corresponding global model versions for model provenance and model quality tracking.\\
\cmidrule(l){2-3}
& {Model replacement trigger} &  Actively monitors the model performance and detects when the degradation in model performance occurs. Replaces degraded models when the degradation persists.\\
\cmidrule(l){2-3}
& {Deployment selector} & Selects and matches the converged global models to suitable client devices to maximise the global model quality for different applications and tasks.\\
\cmidrule(l){2-3}
& {Training configurator} & Enables users to request, configure and deploy FML training processes and models without the need to code or program, using a user-friendly platform.\\

\cmidrule(l){1-3}
\textbf{Model training patterns} & {Multi-task model trainer} & Utilises data from separate but related models on local client devices to improve training efficiency and model quality.\\
\cmidrule(l){2-3}
& {Heterogeneous data handler} & Solves the non-IID and skewed data distribution issues through data volume and data class addition while maintaining the local data privacy.\\
\cmidrule(l){2-3}
& {Incentive registry} & Measures and records the performance and contributions of each client and provides incentives (e.g., cryptocurrencies) to motivate clients' participation. \\

\cmidrule(l){1-3}
\textbf{Model aggregation patterns} & {Asynchronous aggregator} & Performs aggregation asynchronously whenever a model update arrives without waiting for all the model updates every round to reduce aggregation latency.\\
\cmidrule(l){2-3}
& {Decentralised aggregator} & Removes the central server from the system and decentralizes its role to prevent single-point-of-failure and increase reliability.\\
\cmidrule(l){2-3}
& {Hierarchical aggregator} & Adds an edge layer to perform partial aggregation of local models from closely-related client devices to improve model quality and computation efficiency. \\
\cmidrule(l){2-3}
& {Secure aggregator} 
& Adopts secure multiparty computation protocols that manage the model exchange and aggregation security to protect model security. \\

\bottomrule
\end{tabular}
\end{table*}

To design a decision model that elicits the functional and non-functional requirements with the respective patterns, we map the elements of the problem space to the elements of the solution space. The problem space can be presented as a set of functional (FR) or non-functional requirements (NFR), whereas the solution space is a set of patterns targeting to solve the problems. We have adopted the decision model design methodology from~\cite{7516811} and~\cite{9426788} and adopted the notation method from~\cite{7516811} that involves the mapping of requirements and patterns, as shown in Fig.~\ref{fig:Decision model notation}. A single-headed arrow from the pattern to the requirement indicates that the pattern satisfies the requirement. All pattern decisions will have benefits which are indicated by a plus sign (+) and drawbacks which are indicated by a minus sign (-). 

To present patterns combination, a double-headed arrow is used to point from one pattern to another pattern, with the label~\textit{[complements]} for one pattern complementing another, and label~\textit{[alternatives]} for one pattern being an alternative to another pattern. When a pattern complements another pattern it means that the initial pattern is required and the qualities of using the initial pattern also apply to the combination of the patterns. If a system quality is associated with both the initial and the complementary pattern but with different qualifications, the qualification of the complementary pattern overrides the qualification of the initial pattern. A trapezium with a dashed line connected to the respective pattern indicates the conditions or constraints to the adoption of that pattern.

\begin{figure*}
    \centering
    \includegraphics[width=0.8\linewidth]{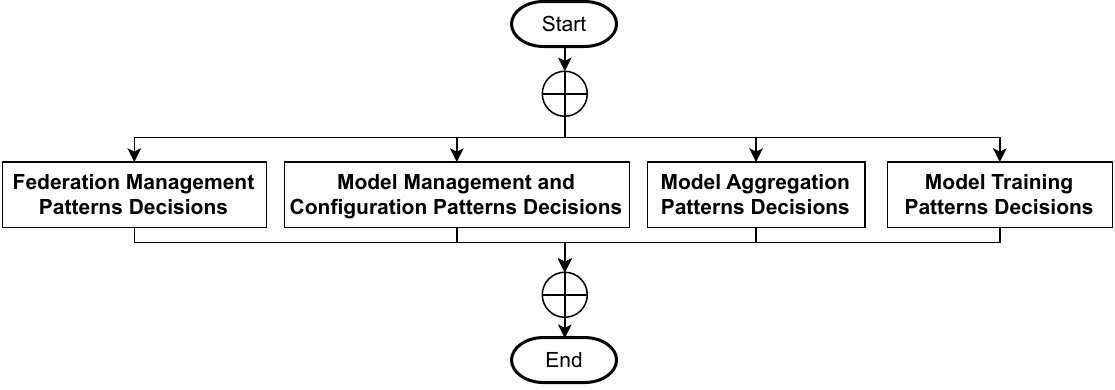}
    \caption{High-level Design Decision Models}
    \label{fig:High_level}
\end{figure*}

We created the elements of the problem and solution spaces based on the categories in the architectural pattern collections that we presented in~\cite{lo2021architectural}, where the summary of the patterns is displayed in Table~\ref{tab:overview}. We have categorised 4 main high-level design decisions of a federated machine learning system and compiled a high-level decision model for federated machine learning system design. As shown in Fig.~\ref{fig:High_level}: (i) Federation management patterns decisions, (2) model management and configuration patterns decisions, (3) model aggregation patterns decisions, and (4) model training patterns decisions. For each lower-level decision within the high-level decisions, the designers need to consider how each quality is positively or negatively affected by another. 

\begin{figure*}[!t]
    \centering
    \includegraphics[width=0.7\linewidth]{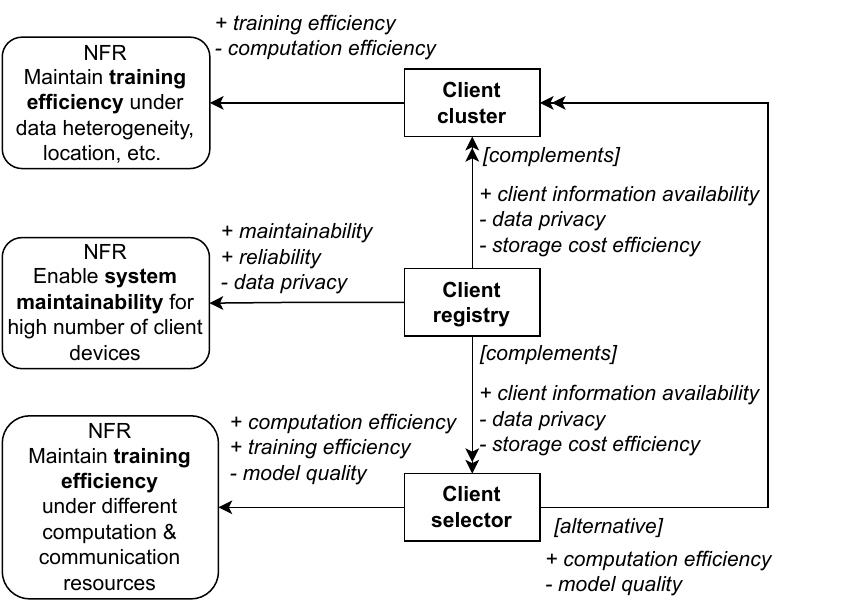}
    \caption{Federation Management Patterns Decision Model}
    \label{fig:Client_management}
\end{figure*}

\subsection{Federation Management Decision Model} \label{Client_Management}

The federation management decision model covers the design decisions that handle device information and the connection between client devices under the federation with the central server, and the selection of client devices for the training process, as shown in Fig.~\ref{fig:Client_management}.

\subsubsection{\textbf{Client cluster}}

The client cluster pattern targets to fulfill the non-functional requirement on \textit{training efficiency} by grouping the client devices with similar characteristics and local models of the same group will be aggregated. In contrast, the extra computational cost is required to access and group the client devices which may reduce the \textit{computation efficiency} of the system. One example is the Iterative Federated Clustering Algorithm (\textit{IFCA})\footnote{https://github.com/jichan3751/ifca}. It is a framework introduced by UC Berkley and Google to cluster client devices based on the loss values of the client's gradient. 

\subsubsection{\textbf{Client registry}}

The \textbf{client registry} pattern could be adopted to enhance the system's \textit{maintainability} and \textit{reliability}. The \textbf{client registry} records the information (smartphones ID, connection uptime/downtime, operating system version, available memory, bandwidth, etc.) of the client devices which are essential for federation management and to schedule when the models are communicated. By entrusting the device information to the central server, the \textit{data privacy} of the clients' information is compromised. The \textbf{client registry} pattern complements the patterns that manage the connection between the central server and the client devices. For instance, the client cluster and client selector patterns both are complemented by the client registry with its \textit{client information availability}. However, this would require the storage of data which induces \textit{privacy} and \textit{storage cost efficiency} issues. One example of this pattern is the \textit{Party Stack} component of \textit{IBM Federated Learning\footnote{https://github.com/IBM/federated-learning-lib}} framework that manages the client parties of IBM federated learning framework with sub-components such as protocol handler, connection, model, local training, and data handler for client devices registration and management.

\subsubsection{\textbf{Client selector}}

The \textbf{client selector} pattern is adopted to actively select the client devices for the training process. This pattern intends to fulfill the non-functional requirement on the \textit{training efficiency} when interacting with client devices that have high differences in their available computation, communication, and memory capacity~\cite{10.1145/3450288,kairouz2019advances}. However, the adoption of \textbf{client selector} excludes a portion of data from clients which may induce low model generalisability, or higher model bias to unseen data and harm the \textit{model quality}~\cite{kairouz2019advances,10.1007/978-3-030-86044-8_6,9686048}. The \textbf{client selector} pattern is an alternative pattern to the \textbf{client cluster} pattern. Both patterns improve the \textit{training efficiency} but the \textbf{client selector} pattern offers better efficiency but may lower \textit{model quality}. One example of this pattern is \textit{IBM's Helios}~\cite{xu2019helios} which has a training consumption profiling function that fully profiles the resource consumption for model training on client devices. Based on the profiling, a resource-aware scheme accelerate local model training on heterogeneous devices and prevent stragglers from delaying the process.

\begin{figure*}[!t]
    \centering
    \includegraphics[width=0.75\linewidth]{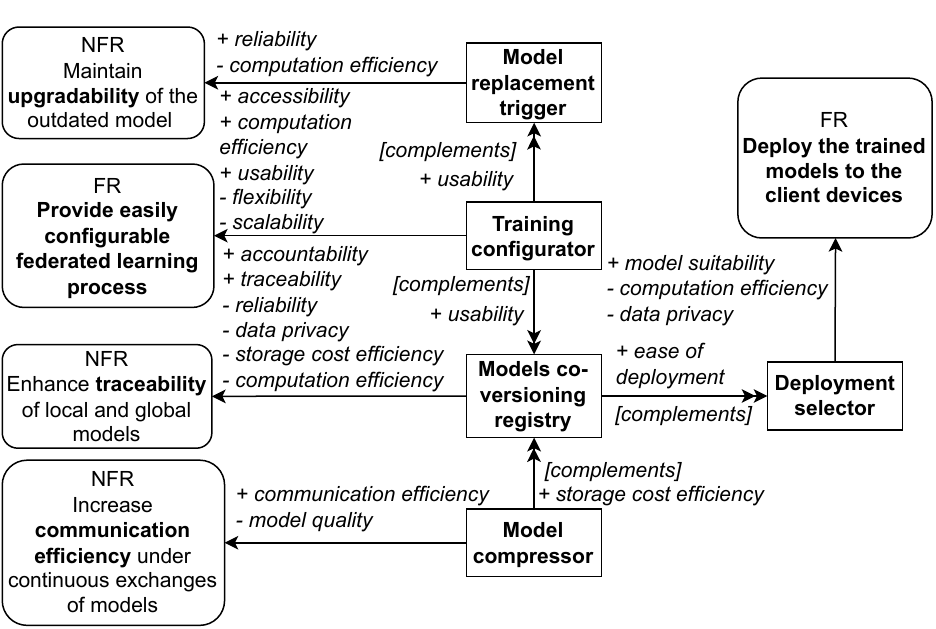}
    \caption{Model Management and Configuration Patterns Decision Model}
    \label{fig:Model_management_Training_Configuration}
\end{figure*}

\subsection{Model Management and Configuration Decision Model} \label{model_management_DM}

\subsubsection{\textbf{Training configurator}}

Fig.~\ref{fig:Model_management_Training_Configuration} shows the model management and configuration decision model. A \textbf{training configurator} provides a user-friendly interface, state-of-the-art practices, and technical support to the system owners to configure the training parameters, client devices management, model management and configuration, and model aggregation mechanisms. This pattern enhances the \textit{accessibility}, \textit{computation efficiency} and \textit{usability} of the federated machine learning system. For instance, \textit{Microsoft Azure Machine Learning Designer}\footnote{\url{https://azure.microsoft.com/en-au/services/machine-learning/designer/}} and the \textit{Amazon SageMaker}\footnote{\url{https://aws.amazon.com/sagemaker/}} are available for centralised or distributed ML systems configuration. However, a preset system may have a relatively lower \textit{flexibility}. Moreover, the system may face \textit{scalability} issues to support more users and devices associated with the expansion of the systems. The \textbf{training configurator} pattern complements the \textbf{model co-versioning registry}, \textbf{model replacement trigger}, and the \textbf{deployment selector} patterns, in terms of \textit{usability}. 

\subsubsection{\textbf{Model co-versioning registry}}

The \textbf{model co-versioning registry} pattern is adopted for model provenance. This approach uses a registry to actively track and record all the model versions and their performance. This effectively increases the \textit{accountability}, and \textit{traceability} of the federated machine learning system. One downside to this approach is the low \textit{storage cost efficiency} due to the requirement to store highly-complex model architecture with the increase in the number of client devices~\cite{lo2021architectural}. Furthermore, users' \textit{data privacy} may be compromised if the registry is managed solely by the central server~\cite{10.1145/3450288,kairouz2019advances}, whereas placing the registry in each client device reduces the devices' \textit{storage cost efficiency} and \textit{computation efficiency}.~\textit{Reliability} issue also occurs when only one party holds the registry. A storage-efficient method using blockchain and smart contracts to track and record only the hashed representations of the data and model versions is mentioned in~\cite{9686048}. In addition, the usage of a combination of decentralised blockchain and database for provenance purposes mentioned in~\cite{9686048,9233457} also resolved the \textit{data privacy} and \textit{trustworthiness} issues. Some examples of this pattern include \textit{DVC}\footnote{https://dvc.org/} which is an online machine learning version control platform built to make models shareable and reproducible, and \textit{Managed MLflow}\footnote{https://databricks.com/product/managed-mlflow} on Databricks that provides chronological model lineage, model versioning, stage transitions, and descriptions.

\subsubsection{\textbf{Model replacement trigger}}

A \textbf{model replacement trigger} pattern monitors the performance of the model that is deployed for real-world usage and when the model's performance (accuracy, precision, etc.) degrades, a request for a new model will be generated to trigger for a new model training task. This maintains the \textit{upgradability} and \textit{reliability} of the systems. However, the continuous monitoring and update of the deployed models will affect the \textit{computation efficiency}. One example is \textit{Microsoft Azure Machine Learning Designer}\footnote{https://azure.microsoft.com/en-au/services/machine-learning/designer/} which is a platform for machine learning pipeline creation that enables models to be retrained on new data.

\subsubsection{\textbf{Deployment selector}}

A \textbf{deployment selector} pattern fulfills the requirement to deploy the trained model to the client devices for real-world usage. This pattern is complemented by the \textbf{training configurator} to increase the model's \textit{ease of deployment}. It deploys converged models based on preset criteria. This is especially crucial in multi-task and multi-model training scenario~\cite{lo2021architectural,kairouz2019advances,10.1145/3450288} where the local data is used to train the models from multiple, different-but-related applications. The \textbf{deployment selector} pattern selects clients that are most suitable to receive the model, according to their specifications, resources, applications, etc. For instance, a smartphone keyboard application provider can deploy the next-world prediction global model to smartphone users that are frequent typists and the text translation global model to smartphone users that use more translation applications for training. This enhances the \textit{model suitability} for each client device's application. Some examples are \textit{Azure Machine Learning}\footnote{https://docs.microsoft.com/en-us/azure/machine-learning/concept-model-management-and-deployment} that supports mass deployment with a step of compute target selection, and \textit{Amazon SageMaker}\footnote{https://docs.aws.amazon.com/sagemaker/latest/dg/multi-model-endpoints.html} which can host multiple models with multi-model endpoints. However, the \textit{data privacy} is compromised as more application and device information are required for the selection criteria and causes lower \textit{computational efficiency}. 

\subsubsection{\textbf{Model compressor}}

A \textbf{model compressor} pattern can be adopted to reduce the data size of the model before being transferred between the two parties. Model pruning and compression can increase the \textit{communication efficiency} but may negatively impact the \textit{model quality} due to the lower model and data precision~\cite{10.1145/3450288,kairouz2019advances}. Using the smartphone use case as an example, the local models are compressed by the app before being sent to the central server, and after aggregation, the updated global model is also compressed before being distributed to the smartphones. One example is Google's structured update and sketched update. The structured update directly learns an update from a restricted space that can be parametrised using a smaller number of variables, whereas sketched update compresses the model before sending it to the central server.

\begin{figure*}[!t]
    \centering
    \includegraphics[width=0.7\linewidth]{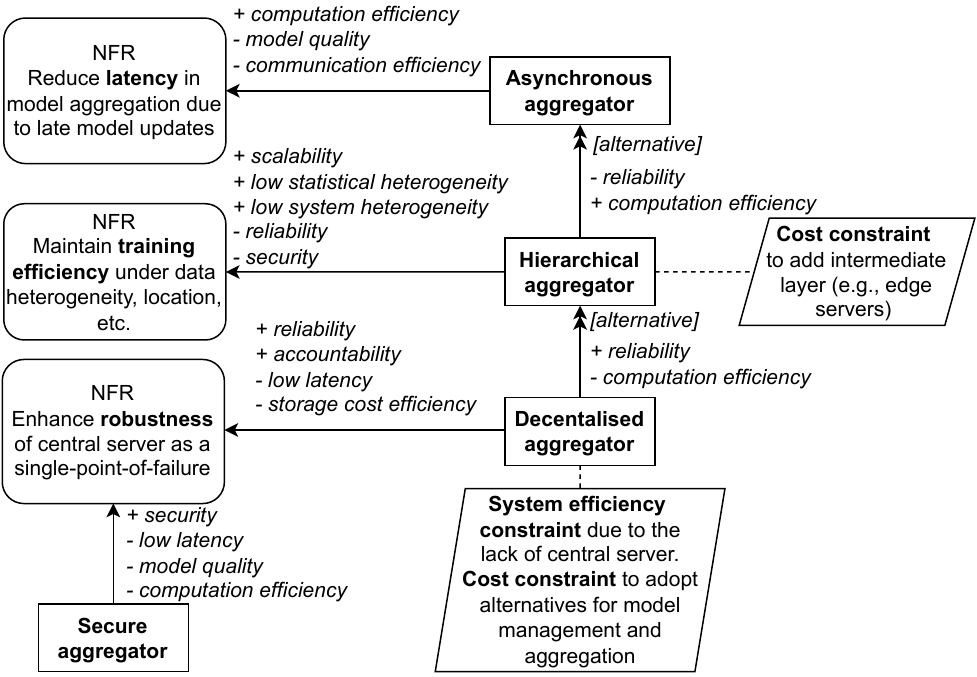}
    \caption{Model Aggregation Patterns Decision Model}
    \label{fig:Model_aggregation}
\end{figure*}

\subsection{Model Aggregation Decision Model} \label{model_aggregation}

\subsubsection{\textbf{Asynchronous aggregator}}

Fig.~\ref{fig:Model_aggregation} shows the model aggregation decision model. The model aggregation time depends on the arrival time of the last local model if synchronous aggregation is performed. To reduce the \textit{latency} in model aggregation and increase the \textit{computation efficiency} of the model aggregation process, the \textbf{asynchronous aggregator} pattern can be adopted. This pattern performs model aggregation whenever a local model update is received, with the currently available global model in the central server. The aggregation \textit{latency} can be reduced which enhances the \textit{computation efficiency}. For instance, the central server aggregates the local model from a smartphone instantly to the latest global model after receiving it. One example is \textit{Asynchronous federated SGD-Vertical Partitioned} (\textit{AFSGD-VP}) \cite{gu2020privacypreserving} algorithm that uses a tree-structured communication scheme to perform asynchronous aggregation. However, a relatively higher number of aggregation rounds may be required and causes lower \textit{communication efficiency}. Some client devices with extremely scarce communication bandwidth may also struggle from being too outdated to join the latest aggregation. Thus, the model produced may be biased. 

\subsubsection{\textbf{Hierarchical aggregator}}

The non-IID (Identically Independently Distributed) data is another main challenge of federated machine learning~\cite{10.1145/3450288,lo2021architectural,9686048}. A \textbf{hierarchical aggregator} pattern uses intermediate layers between the central server and the client devices to resolve the non-IID issue. For instance, edge servers are deployed to cluster and aggregate the smartphones within the same area and perform intermediate model aggregation among these smartphones. By performing an intermediate aggregation, the \textit{statistical} and \textit{system heterogeneity} are improved. The \textit{scalability} is also reduced. One example is \textit{HierFAVG} \cite{liu2019clientedgecloud} which allows multiple edge servers to perform partial model aggregation incrementally from the collected updates. The hierarchical aggregator is an alternative to the asynchronous aggregator that has better \textit{computation efficiency} but lower \textit{reliability}. However, the constraint of this pattern is to have more devices added to the system, which induces higher cost and adds more points-of-failure increase, which compromises \textit{reliability} and \textit{security}.

\subsubsection{\textbf{Decentralised aggregator}}

Another pattern that aims to solve the single-point-of-failure issue is the \textbf{decentralised aggregator} pattern. It removes the central server entirely but it is constrained by low~\textit{computation efficiency} and \textit{cost} of using alternatives to execute model aggregation. These alternatives include using peer-to-peer communication between neighboring client devices~\cite{roy2019braintorrent}, or blockchain and smart contract to manage the models~\cite{8905038,9233457}. For example, instead of sending the local models from each smartphone to a central server owned by the app provider, they share the local models with 10 of their nearest available devices and aggregate their models within the group. This pattern increases the \textit{reliability} and \textit{accountability} in comparison with the centralised federated machine learning approach but it suffers more in terms of \textit{latency} and \textit{storage cost efficiency}, especially due to the peer-to-peer connection and the read/write performance of blockchain. \textit{Swarm Learning}~\cite{warnat2021swarm} is a decentralised aggregator example that utilises edge computing, blockchain-based peer-to-peer networking, and coordination while maintaining confidentiality without the need for a central coordinator. Decentralised aggregator is an alternative to the hierarchical aggregator that increases the \textit{reliability} of the system but lowers \textit{computation efficiency}.

\subsubsection{\textbf{Secure aggregator}}

The central server needs to be robust throughout the training process and hence, a \textbf{secure aggregator} can be adopted. This pattern utilises state-of-the-art security approaches for multiparty computation such as homomorphic encryption to encrypt and decrypt the model before exchanges, or local differential privacy that add noise to the models before exchanges. For instance, \textit{SecAgg} \cite{Bonawitz2017} is a practical protocol by Google for secure aggregation in the federated learning settings. However, differential privacy approaches may reduce the \textit{model quality} due to the noise added to the models. The \textit{computation efficiency} drops and \textit{latency} occurs to perform encryption and decryption for every model update received.

\subsection{Model Training Decision Model} \label{Model_training}

\subsubsection{\textbf{Heterogeneous data handler}}
\begin{figure*}[!t]
    \centering
    \includegraphics[width=0.6\linewidth]{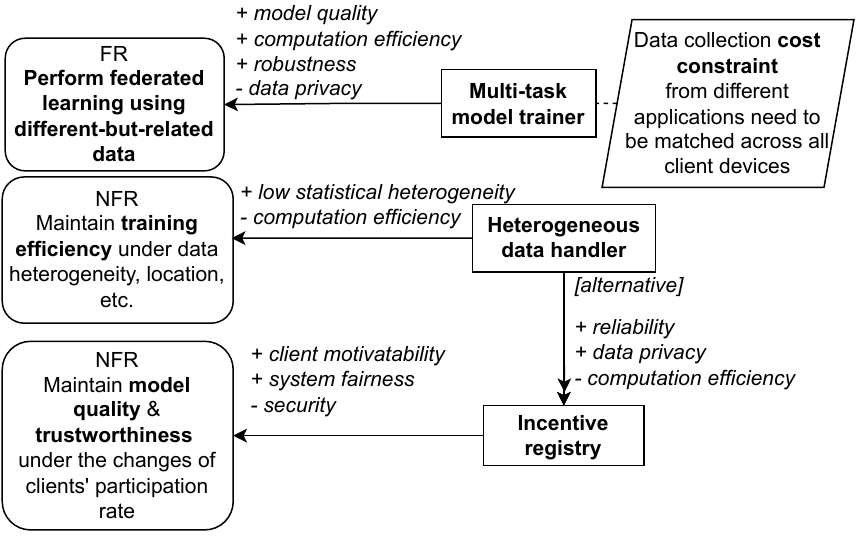}
    \caption{Model Training Patterns Decision Model}
    \label{fig:Model_training}
\end{figure*}

Fig.~\ref{fig:Model_training} shows the model training decision model. Firstly, to solve the \textit{statistical heterogeneity} issue, the \textbf{heterogeneous data handler} pattern can be adopted. The pattern can implement data augmentation for federated machine learning~\cite{jeong2018communication} to generate more data points to create a more balanced dataset, or adopt the federated knowledge distillation method \cite{8904164} that obtains the knowledge from other devices during the distributed training process, without accessing the raw data. For example, the smartphone can use a generative model to create a more balanced dataset (text data with more variety in terms of sentiment) based on the original local data for model training. Apart from solving \textit{statistical heterogeneity}, this pattern also enhances \textit{model quality} in terms of fairness, which serves as an alternative to the \textbf{incentive registry} pattern in terms of \textit{reliability} improvements. For example, federated augmentation (\textit{FAug})~\cite{jeong2018communication} is a data augmentation scheme that utilises a generative adversarial network (GAN) to generate data that takes the tradeoff between privacy leakage and communication overhead into consideration. 

\subsubsection{\textbf{Incentive registry}}
To improve the \textit{model quality} by increasing the participation rate of client devices, the \textbf{incentive registry} can be implemented. By providing a fair amount of reward or compensation to the clients, the overall \textit{client motivatability} is increased and this translates to better model generalisability and hence, better \textit{model quality}. Furthermore, giving rewards according to the clients' contribution can also improve the \textit{system fairness}. Blockchain and smart contract technology are adapted to realise the incentive registry. For example, the app provider can provide exclusive app features or compensation in the form of cryptocurrencies to users that authorised the usage of data and resources to train the local models. However, the provision of rewards may harm the system~\textit{security} as dishonest clients may submit fraudulent results to earn rewards illegally and distort the training process.

\subsubsection{\textbf{Multi-task trainer}}
The \textbf{multi-task trainer} pattern uses local data of different-but-related applications for training to enhance model generalisability for better \textit{model quality} and \textit{robustness}. Furthermore, the training of the model on related or overlapping representations improves the \textit{training efficiency} of the system by reducing the training cost. For instance, a smartphone can use image and text data together to train a multitask model to predict text in an image. However, this pattern is constrained by the requirement to collect and match the data from different applications across all participating client devices to perform multi-task model training. The metadata (features of multiple tasks that are related) of the multiple tasks in the client devices are needed by the central server to create an initial model, which will be challenging in terms of \textit{data privacy}. One example is \textit{MultiModel}\footnote{https://ai.googleblog.com/2017/06/multimodel-multi-task-machine-learning.html} by Google which simultaneously solves several problems spanning multiple domains, including image recognition, translation, and speech recognition.

\section{Architecture Design Validation} \label{evaluation}

In this section, we validate the usability of the decision models by mapping components of existing architectures to the pattern options. We have selected three concrete architectures that are fully maintained and technically supported by top software and web companies: (1) Meta's federated machine learning architecture\footnote{\url{https://engineering.fb.com/2022/06/14/production-engineering/federated-learning-differential-privacy/}}; top hardware company: (2) Intel OpenFL\footnote{{https://openfl.readthedocs.io/en/latest/index.html}}; and top industrial manufacturing company: (3) Siemens Industrial federated learning (IFL)~\cite{hiessl2020industrial}. These architectures by companies that lead industry 4.0 have different design aspects. For instance, Meta focuses on efficiency under differential privacy, Intel focuses on chip-level FML security, and Siemens focuses on fulfilling industrial requirements. We sourced the architectures based on their availability and comprehensiveness at the time of conducting this research.

\subsection{Meta's federated machine learning Architecture}

\begin{figure*}[!t]
\centering
\includegraphics[width=0.7\textwidth]{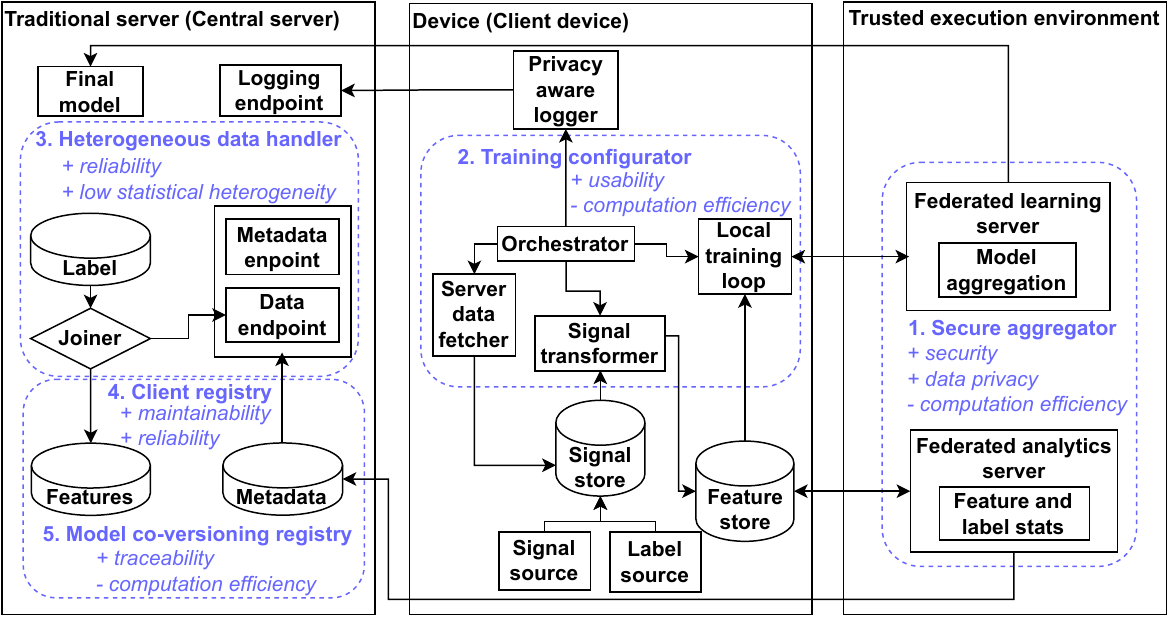}
\caption{Meta's federated machine learning Architecture~\cite{meta}} \label{meta}
\end{figure*}

The mapping of Meta's components onto the decision models is shown in Figure~\ref{meta}.
\begin{itemize}
    \item \textbf{Secure aggregator:} In the trusted execution environment, the federated machine learning server and analytics server applied differential privacy for better \textit{security}.
    
    \item \textbf{Training configurator:} On client devices, the orchestrator, server data fetcher, signal transformer, and local training loop combine to perform: (1) scheduling, (2) device eligibility checking, (2) server-to-device data flow initialisation (3) sample submission control, and (4) logging and performance metric computation~\cite{meta}.
    
    \item \textbf{Heterogeneous data handler:} For better \textit{model quality} in terms of fairness, data/feature augmentations are performed by the joiner component. On the device, the augmentation process is handled by the signal transformer~\cite{meta}.
    
    \item \textbf{Client registry} and \textbf{model co-versioning registry:} Realised by the metadata and feature components under the central server for better \textit{maintainability}, \textit{reliability} and \textit{traceability}.
    
\end{itemize}

\subsection{Intel OpenFL Architecture}
\begin{figure*}[!t]
\centering
\includegraphics[width=0.75\textwidth]{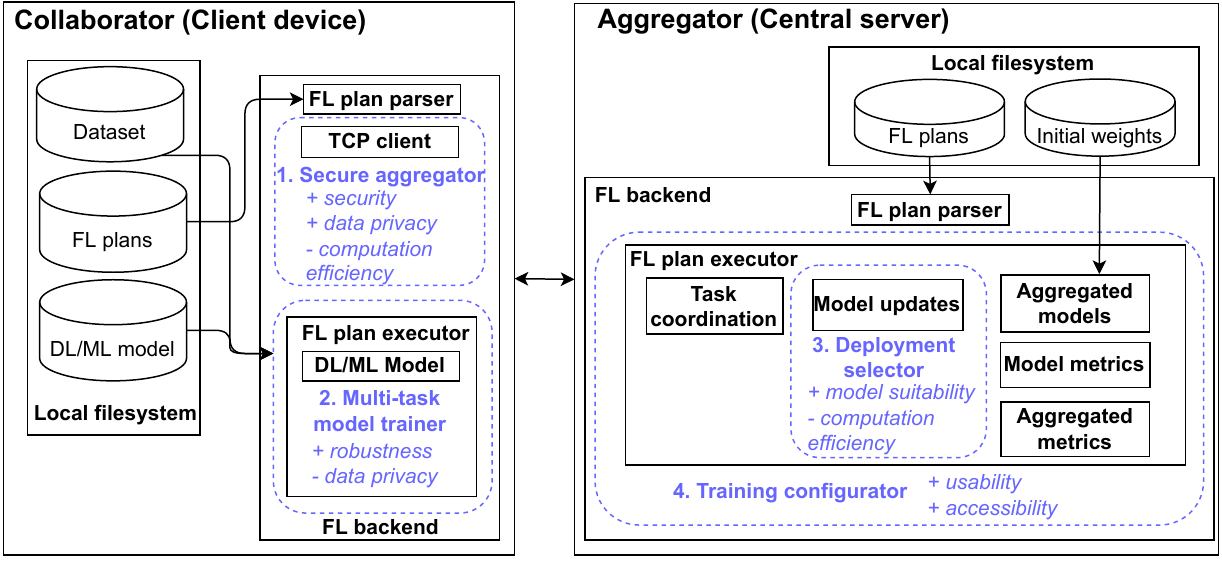}
\caption{Intel OpenFL Architecture~\cite{reina2021openfl}} \label{intel}
\end{figure*}

The mapping of the pattern decisions to Intel's components is shown in Figure~\ref{intel}.

\begin{itemize}
    \item \textbf{Multi-task model trainer:} In the collaborator (client device), the DL/ML model executed by the FL plan executor supports the multi-institutional collaboration for model training to improve the \textit{robustness} of the model~\cite{reina2021openfl}. However, due to the requirement to collect user data, \textit{data privacy} is compromisable.
    
    \item \textbf{Secure aggregator:} The TCP client utilises PKI (Public Key Infrastructure) certificates and mutually authenticated transport layer security (TLS)~\footnote{\url{https://en.wikipedia.org/wiki/Transport_Layer_Security}} connections for better \textit{security} and \textit{data privacy} preservation~\cite{reina2021openfl}.
    
    \item \textbf{Training configurator:} The task coordination component enables the \textit{accessibility} and \textit{usability} of the system. The architecture adapted Trusted Execution Environments (TEEs) to provide hardware mechanisms to execute code with various security properties. It also uses the FL plan to define the collaborator and aggregator settings, such as batch size, IP address, and training rounds, and specifies the remote procedure calls for the given federation tasks~\cite{reina2021openfl}. 
    
    \item \textbf{Deployment selector:} Realised by the model updates component to enhance the \textit{model suitability} for a different collaborator.
\end{itemize}

\subsection{Siemens Industrial federated learning Architecture}

\begin{figure*}[!t]
\centering
\includegraphics[width=0.7\textwidth]{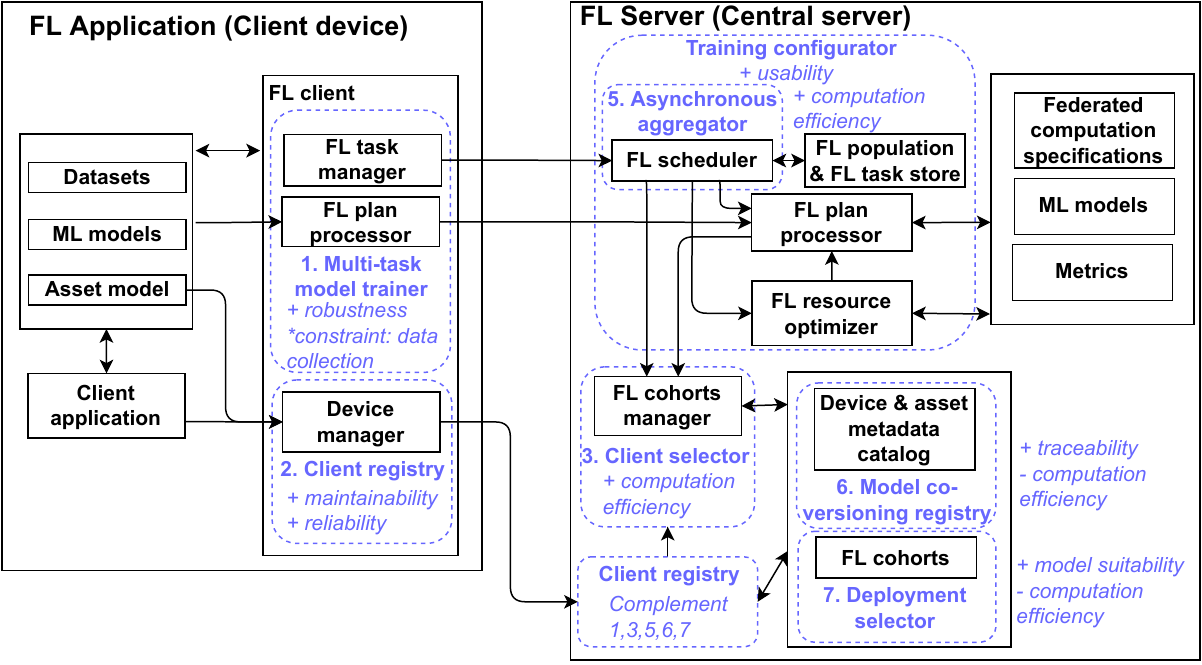}
\caption{Siemens IFL Architecture~\cite{hiessl2020industrial}} \label{siemens}
\end{figure*}
The mapping of Siemens IFL architecture components onto the decision models is shown in Figure~\ref{siemens}. 

\begin{itemize}
    \item \textbf{Multi-task model trainer:} In the FL application (client device), the FL task manager and the FL plan processor realise the \textbf{multi-task model trainer} pattern that identifies learning problems in which multiple FL tasks have in common to increase the \textit{robustness} of the model. However, the adoption is constrained by the client data collection requirement.
    
    \item \textbf{Client registry:} In the FL application (client device), it complements the \textbf{multi-task model trainer} through a device manager to record and manage the client devices. On the server side, it registers the FL clients' organisation, asset data, and data scheme for respective environmental conditions~\cite{hiessl2020industrial}. It also complements all the patterns that require client data, consisting~\textbf{client selector}, \textbf{deployment selector}, \textbf{asynchronous aggregator} and~\textbf{model co-versioning registry} pattern through the utilisation of client data to support their operations. This increases the \textit{maintainability} and \textit{reliability} of the system. However, the pattern induces \textit{computation efficiency} issues. 
    
    \item \textbf{Training configurator:} On the server side, the FL scheduler, FL plan processor, FL resource optimiser, FL population \& FL task store handle the FL plans, training schedules, and resource optimisation to increase the \textit{usability} and \textit{computation efficiency} of the system.
    
    \item \textbf{Client selector}: The FL cohorts manager component acts as the \textit{client selector} to reduce the duration of training or evaluation, which enhances \textit{computation efficiency} and \textit{model quality}. 
    
    \item \textbf{Asynchronous aggregator}: The FL scheduler schedules the FL tasks also improves the \textit{computation efficiency} and reduces aggregation \textit{latency}.
    
    \item \textbf{Model co-versioning registry}: Device \& asset metadata catalog and FL cohorts enhances the model \textit{traceability}.
    
    \item \textbf{Deployment selector:} Uses continuous updates to reevaluate data similarity that is needed to ensure high \textit{model suitability}~\cite{hiessl2020industrial}.
    
\end{itemize}

\subsection{Discussions}

The mappings of the decision models' components and patterns on the existing federated machine learning systems implicated the correctness of the decision models as they mapped the patterns, their benefits, and drawbacks, ideally with the industrial methods and techniques used to address each corresponding requirement through visualisation. However, the is a limitation. The capabilities provided by the pattern have to be there but the components might have different names or also have other responsibilities. Assuming that the architectures are documented properly, it is still possible to find known uses for the patterns but extra efforts are required. 

There are two ways to utilise the decision models in federated machine learning architecture design and validation: (1) Assess the fulfillment of requirements and what are the tradeoffs that might be incurred upon the adaptation of one or more design decisions through visualisation; (2) Extend functionalities of existing architecture to fulfill certain requirements and better identify the tradeoffs that come with the decisions.

\section{Related work} \label{related_work}

A decision model is an approach in software engineering that maps the problems to the solution to guide design decision-making. A well-known approach for creating decision models is Questions-Options-Criteria (QOC)~\cite{maclean1991questions} where the questions represent problems, the options map to solutions, and the criteria are used to determine the options' suitability concerning the questions. Another popular approach from the field of Software Measurement is Goal Questions Metric (GQM)~\cite{caldiera1994goal}. It models the problem according to the goals and questions and provides the metrics to be used for assessing an object and subsequently making decisions to improve it. The Architecture Tradeoff Analysis Method (ATAM)~\cite{kazman2000atam, clements2003evaluating} is a risk-driven approach to evaluating software architectures that helps identify tradeoffs between competing quality attributes, such as performance, security, reliability, and maintainability. 

There is much research work on multiple software engineering and architecture domains that have adopted decision models. For instance, Lewis et al.~\cite{7516811} proposed a decision model for cyber-foraging systems' architectural tactics. Capilla et al.~\cite{10.1007/978-3-030-58923-3_16} highlighted the importance of collaborative decision-making to
produce more accurate and complete design decisions to improve the quality of the architectures. They explored the behavior of software engineering
students as novice software architects in different roles and promote critical design thinking to produce decisions with better quality and architectures. Xu et al.~\cite{9426788} propose a decision model for selecting appropriate patterns for blockchain-based applications. These researchers designed their decision models in extension to the series of patterns or tactics that they have previously published.  

Numerous pieces of research have been conducted on software engineering design decisions for machine learning. For instance, Warnett et al.~\cite{9779697} proposed a series of architectural design decisions for machine learning deployment that covers the decision options, decision drivers, and their relations in the domain of machine learning deployment. Wan et al.~\cite{8812912} studied the effects of machine learning adoptions on software development practices. The work characterizes the differences in various aspects of software engineering and the task involved in machine learning system development and traditional software development. Amershi et al.~\cite{10.1109/ICSE-SEIP.2019.00042} expressed that AI components are more difficult to handle as distinct modules than traditional software components and summarise several best practices to tackle the software engineering challenges in machine learning. Lwakatare et al.~\cite{10.1007/978-3-030-19034-7_14} introduced a taxonomy that depicts machine learning components and their maturity stages of use in the industrial software system by mapping the challenges to the machine learning pipeline stages. Wan et al.~\cite{8945075} studied machine learning design patterns and architectural patterns. Yokohama~\cite{8712157} proposed a set of architectural patterns to improve the operational stability of machine learning systems. A federated machine learning system design was introduced by Bonawitz et. al~\cite{47976}. It focuses on the high-level design of a basic federated machine learning system. In terms of federated machine learning patterns, We compiled and presented a comprehensive and systematic collection of federated machine learning patterns to guide practitioners in developing federated machine learning systems in~\cite{lo2021architectural}. Furthermore, as an extension to the pattern collections, a pattern-oriented reference architecture for federated machine learning was also proposed in~\cite{10.1007/978-3-030-86044-8_6}. Motivated by the aforementioned works, we built decision models for the selection of patterns based on the requirements.

\section{Conclusion and Future Works} \label{conclusion}

This paper presented a set of decision models based on the findings of a systematic literature review that aims to guide academic and industry software architects for federated machine learning system design. The decision models map various functional and non-functional requirements to the patterns, qualified with the benefits and drawbacks to improve the designers' understanding of the effects of the decisions. The decision models have been evaluated in terms of correctness and usefulness through architecture design validations. The mappings of concrete architectures' components to the decision models' options validated the overall usability, correctness, applicability, and comprehensiveness of the decision models. For future works, we plan to expand the decision models by including more patterns, specifically related to the aggregation algorithms and trustworthy AI domain. We also aim to collect more experts' feedback to improve the decision models.

\bibliographystyle{ACM-Reference-Format}
\bibliography{reference}


\begin{thebibliography}{38}


\ifx \showCODEN    \undefined \def \showCODEN     #1{\unskip}     \fi
\ifx \showDOI      \undefined \def \showDOI       #1{#1}\fi
\ifx \showISBNx    \undefined \def \showISBNx     #1{\unskip}     \fi
\ifx \showISBNxiii \undefined \def \showISBNxiii  #1{\unskip}     \fi
\ifx \showISSN     \undefined \def \showISSN      #1{\unskip}     \fi
\ifx \showLCCN     \undefined \def \showLCCN      #1{\unskip}     \fi
\ifx \shownote     \undefined \def \shownote      #1{#1}          \fi
\ifx \showarticletitle \undefined \def \showarticletitle #1{#1}   \fi
\ifx \showURL      \undefined \def \showURL       {\relax}        \fi
\providecommand\bibfield[2]{#2}
\providecommand\bibinfo[2]{#2}
\providecommand\natexlab[1]{#1}
\providecommand\showeprint[2][]{arXiv:#2}

\bibitem[{Ahn} et~al\mbox{.}(2019)]%
        {8904164}
\bibfield{author}{\bibinfo{person}{J. {Ahn}}, \bibinfo{person}{O. {Simeone}},
  {and} \bibinfo{person}{J. {Kang}}.} \bibinfo{year}{2019}\natexlab{}.
\newblock \showarticletitle{Wireless Federated Distillation for Distributed
  Edge Learning with Heterogeneous Data}. In \bibinfo{booktitle}{\emph{PIMRC
  2019}}. \bibinfo{pages}{1--6}.
\newblock


\bibitem[Amershi et~al\mbox{.}(2019)]%
        {10.1109/ICSE-SEIP.2019.00042}
\bibfield{author}{\bibinfo{person}{Saleema Amershi}, \bibinfo{person}{Andrew
  Begel}, \bibinfo{person}{Christian Bird}, \bibinfo{person}{Robert DeLine},
  \bibinfo{person}{Harald Gall}, \bibinfo{person}{Ece Kamar},
  \bibinfo{person}{Nachiappan Nagappan}, \bibinfo{person}{Besmira Nushi}, {and}
  \bibinfo{person}{Thomas Zimmermann}.} \bibinfo{year}{2019}\natexlab{}.
\newblock \showarticletitle{Software Engineering for Machine Learning: A Case
  Study}. In \bibinfo{booktitle}{\emph{Proceedings of the 41st International
  Conference on Software Engineering: Software Engineering in Practice}}
  (Montreal, Quebec, Canada) \emph{(\bibinfo{series}{ICSE-SEIP '19})}.
  \bibinfo{publisher}{IEEE Press}, \bibinfo{pages}{291–300}.
\newblock


\bibitem[{Bao} et~al\mbox{.}(2019)]%
        {8905038}
\bibfield{author}{\bibinfo{person}{X. {Bao}}, \bibinfo{person}{C. {Su}},
  \bibinfo{person}{Y. {Xiong}}, \bibinfo{person}{W. {Huang}}, {and}
  \bibinfo{person}{Y. {Hu}}.} \bibinfo{year}{2019}\natexlab{}.
\newblock \showarticletitle{FLChain: A Blockchain for Auditable Federated
  Learning with Trust and Incentive}. In \bibinfo{booktitle}{\emph{BIGCOM
  '19}}. \bibinfo{pages}{151--159}.
\newblock


\bibitem[Bonawitz et~al\mbox{.}(2017)]%
        {Bonawitz2017}
\bibfield{author}{\bibinfo{person}{Keith Bonawitz}, \bibinfo{person}{Vladimir
  Ivanov}, \bibinfo{person}{Ben Kreuter}, \bibinfo{person}{Antonio Marcedone},
  \bibinfo{person}{H.~Brendan McMahan}, \bibinfo{person}{Sarvar Patel},
  \bibinfo{person}{Daniel Ramage}, \bibinfo{person}{Aaron Segal}, {and}
  \bibinfo{person}{Karn Seth}.} \bibinfo{year}{2017}\natexlab{}.
\newblock \showarticletitle{Practical Secure Aggregation for Privacy-Preserving
  Machine Learning}. In \bibinfo{booktitle}{\emph{Proceedings of the 2017 ACM
  SIGSAC Conference on Computer and Communications Security}} (Dallas, Texas,
  USA) \emph{(\bibinfo{series}{CCS '17})}. \bibinfo{publisher}{Association for
  Computing Machinery}, \bibinfo{address}{New York, NY, USA},
  \bibinfo{pages}{1175–1191}.
\newblock
\showISBNx{9781450349468}


\bibitem[Bonawitz et~al\mbox{.}(2019)]%
        {47976}
\bibfield{author}{\bibinfo{person}{K.~A. Bonawitz}, \bibinfo{person}{Hubert
  Eichner}, \bibinfo{person}{Wolfgang Grieskamp}, \bibinfo{person}{Dzmitry
  Huba}, \bibinfo{person}{Alex Ingerman}, \bibinfo{person}{Vladimir Ivanov},
  \bibinfo{person}{Chloé~M Kiddon}, \bibinfo{person}{Jakub Konečný},
  \bibinfo{person}{Stefano Mazzocchi}, \bibinfo{person}{Brendan McMahan},
  \bibinfo{person}{Timon~Van Overveldt}, \bibinfo{person}{David Petrou},
  \bibinfo{person}{Daniel Ramage}, {and} \bibinfo{person}{Jason Roselander}.}
  \bibinfo{year}{2019}\natexlab{}.
\newblock \showarticletitle{Towards Federated Learning at Scale: System
  Design}. In \bibinfo{booktitle}{\emph{SysML 2019}}.
\newblock
\newblock
\shownote{To appear}.


\bibitem[Caldiera and Rombach(1994)]%
        {caldiera1994goal}
\bibfield{author}{\bibinfo{person}{Victor R Basili1~Gianluigi Caldiera} {and}
  \bibinfo{person}{H~Dieter Rombach}.} \bibinfo{year}{1994}\natexlab{}.
\newblock \showarticletitle{The goal question metric approach}.
\newblock \bibinfo{journal}{\emph{Encyclopedia of software engineering}}
  (\bibinfo{year}{1994}), \bibinfo{pages}{528--532}.
\newblock


\bibitem[Capilla et~al\mbox{.}(2020)]%
        {10.1007/978-3-030-58923-3_16}
\bibfield{author}{\bibinfo{person}{Rafael Capilla}, \bibinfo{person}{Olaf
  Zimmermann}, \bibinfo{person}{Carlos Carrillo}, {and}
  \bibinfo{person}{Hern{\'a}n Astudillo}.} \bibinfo{year}{2020}\natexlab{}.
\newblock \showarticletitle{Teaching Students Software Architecture Decision
  Making}. In \bibinfo{booktitle}{\emph{Software Architecture}},
  \bibfield{editor}{\bibinfo{person}{Anton Jansen}, \bibinfo{person}{Ivano
  Malavolta}, \bibinfo{person}{Henry Muccini}, \bibinfo{person}{Ipek Ozkaya},
  {and} \bibinfo{person}{Olaf Zimmermann}} (Eds.). \bibinfo{publisher}{Springer
  International Publishing}, \bibinfo{address}{Cham},
  \bibinfo{pages}{231--246}.
\newblock
\showISBNx{978-3-030-58923-3}


\bibitem[Clements et~al\mbox{.}(2003)]%
        {clements2003evaluating}
\bibfield{author}{\bibinfo{person}{Paul Clements}, \bibinfo{person}{Rick
  Kazman}, \bibinfo{person}{Mark Klein}, {et~al\mbox{.}}}
  \bibinfo{year}{2003}\natexlab{}.
\newblock \bibinfo{booktitle}{\emph{Evaluating software architectures}}.
\newblock \bibinfo{publisher}{Tsinghua University Press Beijing}.
\newblock


\bibitem[Devlin et~al\mbox{.}(2019)]%
        {devlin2019bert}
\bibfield{author}{\bibinfo{person}{Jacob Devlin}, \bibinfo{person}{Ming-Wei
  Chang}, \bibinfo{person}{Kenton Lee}, {and} \bibinfo{person}{Kristina
  Toutanova}.} \bibinfo{year}{2019}\natexlab{}.
\newblock \bibinfo{title}{BERT: Pre-training of Deep Bidirectional Transformers
  for Language Understanding}.
\newblock
\newblock
\showeprint[arxiv]{1810.04805}~[cs.CL]


\bibitem[Gu et~al\mbox{.}(2020)]%
        {gu2020privacypreserving}
\bibfield{author}{\bibinfo{person}{Bin Gu}, \bibinfo{person}{An Xu},
  \bibinfo{person}{Zhouyuan Huo}, \bibinfo{person}{Cheng Deng}, {and}
  \bibinfo{person}{Heng Huang}.} \bibinfo{year}{2020}\natexlab{}.
\newblock \bibinfo{title}{Privacy-Preserving Asynchronous Federated Learning
  Algorithms for Multi-Party Vertically Collaborative Learning}.
\newblock
\newblock


\bibitem[Hiessl et~al\mbox{.}(2020)]%
        {hiessl2020industrial}
\bibfield{author}{\bibinfo{person}{Thomas Hiessl}, \bibinfo{person}{Daniel
  Schall}, \bibinfo{person}{Jana Kemnitz}, {and} \bibinfo{person}{Stefan
  Schulte}.} \bibinfo{year}{2020}\natexlab{}.
\newblock \showarticletitle{Industrial federated learning--requirements and
  system design}. In \bibinfo{booktitle}{\emph{International Conference on
  Practical Applications of Agents and Multi-Agent Systems}}. Springer,
  \bibinfo{pages}{42--53}.
\newblock


\bibitem[Jeong et~al\mbox{.}(2018)]%
        {jeong2018communication}
\bibfield{author}{\bibinfo{person}{Eunjeong Jeong}, \bibinfo{person}{Seungeun
  Oh}, \bibinfo{person}{Hyesung Kim}, \bibinfo{person}{Jihong Park},
  \bibinfo{person}{Mehdi Bennis}, {and} \bibinfo{person}{Seong-Lyun Kim}.}
  \bibinfo{year}{2018}\natexlab{}.
\newblock \showarticletitle{Communication-efficient on-device machine learning:
  Federated distillation and augmentation under non-iid private data}.
\newblock \bibinfo{journal}{\emph{arXiv preprint arXiv:1811.11479}}
  (\bibinfo{year}{2018}).
\newblock


\bibitem[Jobin et~al\mbox{.}(2019)]%
        {jobin2019global}
\bibfield{author}{\bibinfo{person}{Anna Jobin}, \bibinfo{person}{Marcello
  Ienca}, {and} \bibinfo{person}{Effy Vayena}.}
  \bibinfo{year}{2019}\natexlab{}.
\newblock \showarticletitle{The global landscape of {AI} ethics guidelines}.
\newblock \bibinfo{journal}{\emph{Nature Machine Intelligence}}
  \bibinfo{volume}{1}, \bibinfo{number}{9} (\bibinfo{year}{2019}),
  \bibinfo{pages}{389--399}.
\newblock


\bibitem[Kairouz et~al\mbox{.}(2019)]%
        {kairouz2019advances}
\bibfield{author}{\bibinfo{person}{Peter Kairouz}, \bibinfo{person}{H~Brendan
  McMahan}, \bibinfo{person}{Brendan Avent}, \bibinfo{person}{Aur{\'e}lien
  Bellet}, \bibinfo{person}{Mehdi Bennis}, \bibinfo{person}{Arjun~Nitin
  Bhagoji}, \bibinfo{person}{Keith Bonawitz}, \bibinfo{person}{Zachary
  Charles}, \bibinfo{person}{Graham Cormode}, \bibinfo{person}{Rachel
  Cummings}, {et~al\mbox{.}}} \bibinfo{year}{2019}\natexlab{}.
\newblock \showarticletitle{Advances and open problems in federated learning}.
\newblock \bibinfo{journal}{\emph{arXiv preprint arXiv:1912.04977}}
  (\bibinfo{year}{2019}).
\newblock


\bibitem[Kazman et~al\mbox{.}(2000)]%
        {kazman2000atam}
\bibfield{author}{\bibinfo{person}{Rick Kazman}, \bibinfo{person}{Mark Klein},
  {and} \bibinfo{person}{Paul Clements}.} \bibinfo{year}{2000}\natexlab{}.
\newblock \bibinfo{booktitle}{\emph{ATAM: Method for architecture evaluation}}.
\newblock \bibinfo{type}{{T}echnical {R}eport}.
  \bibinfo{institution}{Carnegie-Mellon Univ Pittsburgh PA Software Engineering
  Inst}.
\newblock


\bibitem[Lewis et~al\mbox{.}(2016)]%
        {7516811}
\bibfield{author}{\bibinfo{person}{Grace~A. Lewis}, \bibinfo{person}{Patricia
  Lago}, {and} \bibinfo{person}{Paris Avgeriou}.}
  \bibinfo{year}{2016}\natexlab{}.
\newblock \showarticletitle{A Decision Model for Cyber-Foraging Systems}. In
  \bibinfo{booktitle}{\emph{2016 13th Working IEEE/IFIP Conference on Software
  Architecture (WICSA)}}. \bibinfo{pages}{51--60}.
\newblock


\bibitem[Liu et~al\mbox{.}(2019)]%
        {liu2019clientedgecloud}
\bibfield{author}{\bibinfo{person}{Lumin Liu}, \bibinfo{person}{Jun Zhang},
  \bibinfo{person}{S.~H. Song}, {and} \bibinfo{person}{Khaled~B. Letaief}.}
  \bibinfo{year}{2019}\natexlab{}.
\newblock \bibinfo{title}{Client-Edge-Cloud Hierarchical Federated Learning}.
\newblock
\newblock
\showeprint[arxiv]{1905.06641}~[cs.NI]


\bibitem[Lo et~al\mbox{.}(2019)]%
        {s19204354}
\bibfield{author}{\bibinfo{person}{Sin~Kit Lo}, \bibinfo{person}{Chee~Sun
  Liew}, \bibinfo{person}{Kok~Soon Tey}, {and} \bibinfo{person}{Saad
  Mekhilef}.} \bibinfo{year}{2019}\natexlab{}.
\newblock \showarticletitle{An Interoperable Component-Based Architecture for
  Data-Driven IoT System}.
\newblock \bibinfo{journal}{\emph{Sensors}} \bibinfo{volume}{19},
  \bibinfo{number}{20} (\bibinfo{year}{2019}).
\newblock
\showISSN{1424-8220}


\bibitem[Lo et~al\mbox{.}(2022a)]%
        {9686048}
\bibfield{author}{\bibinfo{person}{Sin~Kit Lo}, \bibinfo{person}{Yue Liu},
  \bibinfo{person}{Qinghua Lu}, \bibinfo{person}{Chen Wang},
  \bibinfo{person}{Xiwei Xu}, \bibinfo{person}{Hye-Young Paik}, {and}
  \bibinfo{person}{Liming Zhu}.} \bibinfo{year}{2022}\natexlab{a}.
\newblock \showarticletitle{Towards Trustworthy AI: Blockchain-based
  Architecture Design for Accountability and Fairness of Federated Learning
  Systems}.
\newblock \bibinfo{journal}{\emph{IEEE Internet of Things Journal}}
  (\bibinfo{year}{2022}), \bibinfo{pages}{1--1}.
\newblock


\bibitem[Lo et~al\mbox{.}(2021a)]%
        {10.1007/978-3-030-86044-8_6}
\bibfield{author}{\bibinfo{person}{Sin~Kit Lo}, \bibinfo{person}{Qinghua Lu},
  \bibinfo{person}{Hye-Young Paik}, {and} \bibinfo{person}{Liming Zhu}.}
  \bibinfo{year}{2021}\natexlab{a}.
\newblock \showarticletitle{{FLRA}: A Reference Architecture for Federated
  Learning Systems}. In \bibinfo{booktitle}{\emph{Software Architecture}}.
  \bibinfo{publisher}{Springer International Publishing},
  \bibinfo{pages}{83--98}.
\newblock
\showISBNx{978-3-030-86044-8}


\bibitem[Lo et~al\mbox{.}(2021b)]%
        {10.1145/3450288}
\bibfield{author}{\bibinfo{person}{Sin~Kit Lo}, \bibinfo{person}{Qinghua Lu},
  \bibinfo{person}{Chen Wang}, \bibinfo{person}{Hye-Young Paik}, {and}
  \bibinfo{person}{Liming Zhu}.} \bibinfo{year}{2021}\natexlab{b}.
\newblock \showarticletitle{A Systematic Literature Review on Federated Machine
  Learning: From a Software Engineering Perspective}.
\newblock \bibinfo{journal}{\emph{ACM Comput. Surv.}} \bibinfo{volume}{54},
  \bibinfo{number}{5}, Article \bibinfo{articleno}{95} (\bibinfo{date}{May}
  \bibinfo{year}{2021}), \bibinfo{numpages}{39}~pages.
\newblock
\showISSN{0360-0300}


\bibitem[Lo et~al\mbox{.}(2022b)]%
        {lo2021architectural}
\bibfield{author}{\bibinfo{person}{Sin~Kit Lo}, \bibinfo{person}{Qinghua Lu},
  \bibinfo{person}{Liming Zhu}, \bibinfo{person}{Hye-Young Paik},
  \bibinfo{person}{Xiwei Xu}, {and} \bibinfo{person}{Chen Wang}.}
  \bibinfo{year}{2022}\natexlab{b}.
\newblock \showarticletitle{Architectural patterns for the design of federated
  learning systems}.
\newblock \bibinfo{journal}{\emph{Journal of Systems and Software}}
  \bibinfo{volume}{191} (\bibinfo{year}{2022}), \bibinfo{pages}{111357}.
\newblock
\showISSN{0164-1212}


\bibitem[Lwakatare et~al\mbox{.}(2019)]%
        {10.1007/978-3-030-19034-7_14}
\bibfield{author}{\bibinfo{person}{Lucy~Ellen Lwakatare},
  \bibinfo{person}{Aiswarya Raj}, \bibinfo{person}{Jan Bosch},
  \bibinfo{person}{Helena~Holmstr\"om Olsson}, {and} \bibinfo{person}{Ivica
  Crnkovic}.} \bibinfo{year}{2019}\natexlab{}.
\newblock \bibinfo{booktitle}{\emph{A Taxonomy of Software Engineering
  Challenges for Machine Learning Systems: An Empirical Investigation}}.
\newblock \bibinfo{pages}{227--243}.
\newblock


\bibitem[MacLean et~al\mbox{.}(1991)]%
        {maclean1991questions}
\bibfield{author}{\bibinfo{person}{Allan MacLean}, \bibinfo{person}{Richard~M
  Young}, \bibinfo{person}{Victoria~ME Bellotti}, {and}
  \bibinfo{person}{Thomas~P Moran}.} \bibinfo{year}{1991}\natexlab{}.
\newblock \showarticletitle{Questions, options, and criteria: Elements of
  design space analysis}.
\newblock \bibinfo{journal}{\emph{Human--computer interaction}}
  \bibinfo{volume}{6}, \bibinfo{number}{3-4} (\bibinfo{year}{1991}),
  \bibinfo{pages}{201--250}.
\newblock


\bibitem[McMahan et~al\mbox{.}(2017)]%
        {mcmahan2017communicationefficient}
\bibfield{author}{\bibinfo{person}{H.~Brendan McMahan}, \bibinfo{person}{Eider
  Moore}, \bibinfo{person}{Daniel Ramage}, \bibinfo{person}{Seth Hampson},
  {and} \bibinfo{person}{Blaise~Agüera y Arcas}.}
  \bibinfo{year}{2017}\natexlab{}.
\newblock \bibinfo{title}{Communication-Efficient Learning of Deep Networks
  from Decentralized Data}.
\newblock
\newblock
\showeprint[arxiv]{1602.05629}~[cs.LG]


\bibitem[Reina et~al\mbox{.}(2021)]%
        {reina2021openfl}
\bibfield{author}{\bibinfo{person}{G~Anthony Reina}, \bibinfo{person}{Alexey
  Gruzdev}, \bibinfo{person}{Patrick Foley}, \bibinfo{person}{Olga
  Perepelkina}, \bibinfo{person}{Mansi Sharma}, \bibinfo{person}{Igor
  Davidyuk}, \bibinfo{person}{Ilya Trushkin}, \bibinfo{person}{Maksim
  Radionov}, \bibinfo{person}{Aleksandr Mokrov}, \bibinfo{person}{Dmitry
  Agapov}, {et~al\mbox{.}}} \bibinfo{year}{2021}\natexlab{}.
\newblock \showarticletitle{OpenFL: An open-source framework for Federated
  Learning}.
\newblock \bibinfo{journal}{\emph{arXiv preprint arXiv:2105.06413}}
  (\bibinfo{year}{2021}).
\newblock


\bibitem[Roy et~al\mbox{.}(2019)]%
        {roy2019braintorrent}
\bibfield{author}{\bibinfo{person}{Abhijit~Guha Roy}, \bibinfo{person}{Shayan
  Siddiqui}, \bibinfo{person}{Sebastian Pölsterl}, \bibinfo{person}{Nassir
  Navab}, {and} \bibinfo{person}{Christian Wachinger}.}
  \bibinfo{year}{2019}\natexlab{}.
\newblock \bibinfo{title}{BrainTorrent: A Peer-to-Peer Environment for
  Decentralized Federated Learning}.
\newblock
\newblock
\showeprint[arxiv]{1905.06731}~[cs.LG]


\bibitem[Stojkovic et~al\mbox{.}(2022)]%
        {meta}
\bibfield{author}{\bibinfo{person}{Branislav Stojkovic},
  \bibinfo{person}{Jonathan Woodbridge}, \bibinfo{person}{Zhihan Fang},
  \bibinfo{person}{Jerry Cai}, \bibinfo{person}{Andrey Petrov},
  \bibinfo{person}{Sathya Iyer}, \bibinfo{person}{Daoyu Huang},
  \bibinfo{person}{Patrick Yau}, \bibinfo{person}{Arvind~Sastha Kumar},
  \bibinfo{person}{Hitesh Jawa}, {and} \bibinfo{person}{Anamita Guha}.}
  \bibinfo{year}{2022}\natexlab{}.
\newblock \bibinfo{title}{Applied Federated Learning: Architectural Design for
  Robust and Efficient Learning in Privacy Aware Settings}.
\newblock
\newblock
\urldef\tempurl%
\url{https://doi.org/10.48550/ARXIV.2206.00807}
\showDOI{\tempurl}


\bibitem[Thiebes et~al\mbox{.}(2020)]%
        {TAI2020}
\bibfield{author}{\bibinfo{person}{Scott Thiebes}, \bibinfo{person}{Sebastian
  Lins}, {and} \bibinfo{person}{Ali Sunyaev}.} \bibinfo{year}{2020}\natexlab{}.
\newblock \showarticletitle{Trustworthy artificial intelligence}.
\newblock \bibinfo{journal}{\emph{Electronic Markets}} (\bibinfo{year}{2020}).
\newblock
\showISSN{1422-8890}


\bibitem[{Wan} et~al\mbox{.}(2019)]%
        {8812912}
\bibfield{author}{\bibinfo{person}{Z. {Wan}}, \bibinfo{person}{X. {Xia}},
  \bibinfo{person}{D. {Lo}}, {and} \bibinfo{person}{G.~C. {Murphy}}.}
  \bibinfo{year}{2019}\natexlab{}.
\newblock \showarticletitle{How does Machine Learning Change Software
  Development Practices?}
\newblock \bibinfo{journal}{\emph{IEEE Trans. Softw. Eng.}}
  (\bibinfo{year}{2019}), \bibinfo{pages}{1}.
\newblock


\bibitem[Warnat-Herresthal et~al\mbox{.}(2021)]%
        {warnat2021swarm}
\bibfield{author}{\bibinfo{person}{Stefanie Warnat-Herresthal},
  \bibinfo{person}{Hartmut Schultze},
  \bibinfo{person}{Krishnaprasad~Lingadahalli Shastry},
  \bibinfo{person}{Sathyanarayanan Manamohan}, \bibinfo{person}{Saikat
  Mukherjee}, \bibinfo{person}{Vishesh Garg}, \bibinfo{person}{Ravi
  Sarveswara}, \bibinfo{person}{Kristian H\"andler}, \bibinfo{person}{Peter
  Pickkers}, \bibinfo{person}{N~Ahmad Aziz}, {et~al\mbox{.}}}
  \bibinfo{year}{2021}\natexlab{}.
\newblock \showarticletitle{Swarm Learning for decentralized and confidential
  clinical machine learning}.
\newblock \bibinfo{journal}{\emph{Nature}} \bibinfo{volume}{594},
  \bibinfo{number}{7862} (\bibinfo{year}{2021}), \bibinfo{pages}{265--270}.
\newblock


\bibitem[Warnett and Zdun(2022)]%
        {9779697}
\bibfield{author}{\bibinfo{person}{Stephen~John Warnett} {and}
  \bibinfo{person}{Uwe Zdun}.} \bibinfo{year}{2022}\natexlab{}.
\newblock \showarticletitle{Architectural Design Decisions for Machine Learning
  Deployment}. In \bibinfo{booktitle}{\emph{2022 IEEE 19th International
  Conference on Software Architecture (ICSA)}}. \bibinfo{pages}{90--100}.
\newblock


\bibitem[{Washizaki} et~al\mbox{.}(2019)]%
        {8945075}
\bibfield{author}{\bibinfo{person}{H. {Washizaki}}, \bibinfo{person}{H.
  {Uchida}}, \bibinfo{person}{F. {Khomh}}, {and} \bibinfo{person}{Y.
  {Gu\'eh\'eneuc}}.} \bibinfo{year}{2019}\natexlab{}.
\newblock


\bibitem[Xu et~al\mbox{.}(2021a)]%
        {9426788}
\bibfield{author}{\bibinfo{person}{Xiwei Xu}, \bibinfo{person}{H.M.N.
  Dilum~Bandara}, \bibinfo{person}{Qinghua Lu}, \bibinfo{person}{Ingo Weber},
  \bibinfo{person}{Len Bass}, {and} \bibinfo{person}{Liming Zhu}.}
  \bibinfo{year}{2021}\natexlab{a}.
\newblock \showarticletitle{A Decision Model for Choosing Patterns in
  Blockchain-Based Applications}. In \bibinfo{booktitle}{\emph{2021 IEEE 18th
  International Conference on Software Architecture (ICSA)}}.
  \bibinfo{pages}{47--57}.
\newblock


\bibitem[Xu et~al\mbox{.}(2021b)]%
        {xu2019helios}
\bibfield{author}{\bibinfo{person}{Zirui Xu}, \bibinfo{person}{Fuxun Yu},
  \bibinfo{person}{Jinjun Xiong}, {and} \bibinfo{person}{Xiang Chen}.}
  \bibinfo{year}{2021}\natexlab{b}.
\newblock \showarticletitle{Helios: Heterogeneity-Aware Federated Learning with
  Dynamically Balanced Collaboration}. In \bibinfo{booktitle}{\emph{2021 58th
  ACM/IEEE Design Automation Conference (DAC)}}. \bibinfo{pages}{997--1002}.
\newblock
\urldef\tempurl%
\url{https://doi.org/10.1109/DAC18074.2021.9586241}
\showDOI{\tempurl}


\bibitem[Yang et~al\mbox{.}(2019)]%
        {10.1145/3298981}
\bibfield{author}{\bibinfo{person}{Qiang Yang}, \bibinfo{person}{Yang Liu},
  \bibinfo{person}{Tianjian Chen}, {and} \bibinfo{person}{Yongxin Tong}.}
  \bibinfo{year}{2019}\natexlab{}.
\newblock \showarticletitle{Federated Machine Learning: Concept and
  Applications}.
\newblock \bibinfo{journal}{\emph{ACM Trans. Intell. Syst. Technol.}}
  \bibinfo{volume}{10}, \bibinfo{number}{2}, Article \bibinfo{articleno}{12}
  (\bibinfo{date}{jan} \bibinfo{year}{2019}), \bibinfo{numpages}{19}~pages.
\newblock
\showISSN{2157-6904}


\bibitem[{Yokoyama}(2019)]%
        {8712157}
\bibfield{author}{\bibinfo{person}{H. {Yokoyama}}.}
  \bibinfo{year}{2019}\natexlab{}.
\newblock \bibinfo{booktitle}{\emph{Machine Learning System Architectural
  Pattern for Improving Operational Stability}}.
\newblock \bibinfo{pages}{267--274}.
\newblock


\bibitem[{Zhang} et~al\mbox{.}(2020)]%
        {9233457}
\bibfield{author}{\bibinfo{person}{W. {Zhang}}, \bibinfo{person}{Q. {Lu}},
  \bibinfo{person}{Q. {Yu}}, \bibinfo{person}{Z. {Li}}, \bibinfo{person}{Y.
  {Liu}}, \bibinfo{person}{S.~K. {Lo}}, \bibinfo{person}{S. {Chen}},
  \bibinfo{person}{X. {Xu}}, {and} \bibinfo{person}{L. {Zhu}}.}
  \bibinfo{year}{2020}\natexlab{}.
\newblock \showarticletitle{Blockchain-based Federated Learning for Device
  Failure Detection in Industrial IoT}.
\newblock \bibinfo{journal}{\emph{IEEE Internet Things J.}}
  (\bibinfo{year}{2020}), \bibinfo{pages}{1--12}.
\newblock


\end{thebibliography}

\end{document}